# Structured Context Engineering for File-Native Agentic Systems

Evaluating Schema Accuracy, Format Effectiveness, and Multi-File Navigation at Scale

Damon McMillan

*HxAI Australia*

**Abstract**

Large Language Model agents increasingly operate external systems through programmatic interfaces, yet practitioners lack empirical guidance on how to structure the context these agents consume. Using SQL generation as a proxy for programmatic agent operations, we present a systematic study of context engineering for structured data, comprising 9,649 experiments across 11 models, 4 formats (YAML, Markdown, JSON, Token-Oriented Object Notation [TOON]), and schemas ranging from 10 to 10,000 tables.

Our findings challenge common assumptions. First, architecture choice is model-dependent: file-based context retrieval improves accuracy for frontier-tier models (Claude, GPT, Gemini; +2.7%, p=0.029) but shows mixed results for open source models (aggregate -7.7%, p<0.001), with deficits varying substantially by model. Second, format does not significantly affect aggregate accuracy (chi-squared=2.45, p=0.484), though individual models, particularly open source, exhibit format-specific sensitivities. Third, model capability is the dominant factor, with a 21 percentage point accuracy gap between frontier and open source tiers that dwarfs any format or architecture effect. Fourth, file-native agents scale to 10,000 tables through domain-partitioned schemas while maintaining high navigation accuracy. Fifth, file size does not predict runtime efficiency: compact or novel formats can incur a token overhead driven by grep output density and pattern unfamiliarity, with the magnitude depending on model capability.

These findings provide practitioners with evidence-based guidance for deploying LLM agents on structured systems, demonstrating that architectural decisions should be tailored to model capability rather than assuming universal best practices.

## 1 Introduction

The emergence of Large Language Model (LLM) agents capable of operating external systems has created a fundamental question in AI systems design: how should agents understand the systems they actuate? Unlike information retrieval or question answering, agentic tasks require the LLM to comprehend schemas, relationships, constraints, and semantic meaning well enough to generate correct operations. This paper investigates structured context engineering for file-native agentic systems: the practice of organising persistent system knowledge to optimise agent performance when agents retrieve context through native file operations.

### 1.1 The Rise of File-Native Agent Context

A significant shift is occurring in how developers provide context to LLM agents. Rather than relying solely on retrieval-augmented approaches or embedding context directly in prompts, practitioners are increasingly adopting file-based semantic layers: structured documents that agents read through native file tools such as grep and read operations.

This pattern has emerged organically across the industry. CLAUDE.md and AGENTS.md files describe project conventions and architecture for coding agents. The llms.txt standard provides structured website descriptions optimised for LLM consumption. Cursor Rules and similar IDE configurations structure context for code-focused agents. Schema files in YAML, JSON, or Markdown describe database structures and API specifications for data agents.

## 1.2 Research Questions

This paper investigates five research questions:

**Table 1: Research Questions**

| RQ | Question |
|---|---|
| R1 | Does file-native context engineering provide better accuracy than prompt engineering? |
| R2 | Does format affect accuracy for file-native agents? |
| R3 | How does model and model tier affect context engineering effectiveness? |
| R4 | How does schema scale affect file-native agent effectiveness? |
| R5 | Does format affect efficiency (tokens, cost, tool calls)? |

## 1.3 Contributions

This paper makes the following contributions: (1) a systematic study with 9,649 experiments across 11 models, 4 formats, 2 architectures, and schemas from 10 to 10,000 tables; (2) demonstration that file-native architecture benefits frontier models but shows mixed results for open source models; (3) finding no significant aggregate format effect while identifying model-specific sensitivities; (4) validation of high navigation accuracy at 10,000 tables using partitioning; (5) identification of the 'grep tax' phenomenon where compact or novel formats consume more runtime tokens due to grep output density and pattern unfamiliarity, with documented failure modes.

## 2 Related Work

The evolution from single-turn LLM interactions to agentic systems has created new paradigms for how models interact with external systems. The ReAct framework [1] established the reasoning-plus-acting paradigm where models interleave chain-of-thought reasoning with tool invocations. Toolformer [11] demonstrated that language models can learn to use tools autonomously, while Gorilla [16] showed that LLMs can be connected with massive APIs for real-world actuation. ToolLLM [23] and ToolACE [25] have further advanced LLM function calling across thousands of real-world APIs. The Model Context Protocol and similar standards have formalised how agents access external resources [2]. Agent evaluation has advanced through benchmarks such as SWE-bench [14] for software engineering tasks and AgentBench [15] for evaluating LLMs across diverse agent environments.

Text-to-SQL translation has advanced significantly with large language models. The Spider benchmark [3] established cross-domain evaluation standards, followed by BIRD [17] which introduced large-scale, real-world databases with noisy schemas. DIN-SQL [4] achieved strong results by decomposing complex queries, while MAC-SQL [18] demonstrated multi-agent collaborative approaches to text-to-SQL. Gao et al. [5] provided a comprehensive benchmark evaluation of LLM-based text-to-SQL approaches. Recent work has also questioned whether explicit schema linking remains necessary with capable models [24]. Our work differs by focusing on context engineering rather than prompt techniques, investigating how schema format and retrieval architecture affect generation accuracy.

Mohsenimofidi et al. [7] conducted the first empirical study of AI configuration files across 466 repositories, identifying a research gap: how content, structure, and style affect agent behaviour remains unevaluated. Our contribution directly addresses this gap.

Context engineering has recently been formalised as a discipline distinct from prompt engineering. A comprehensive survey [12] covering over 1,400 papers establishes context engineering as the systematic optimisation of information payloads for LLMs, encompassing retrieval, processing, and management. Chain-of-thought prompting [13] demonstrated that explicit reasoning steps improve performance on complex tasks, a finding relevant to our complexity tier analysis. StructGPT [21] proposed an iterative reading-then-reasoning framework for structured data including databases and knowledge graphs. Notably, the 'lost in the middle' phenomenon [6] demonstrates that LLMs struggle to use information positioned in context middles, motivating selective retrieval approaches like our file-native architecture.

Recent work has examined how serialisation format affects LLM comprehension of structured data. Sui et al. [19] benchmarked table serialisation formats (HTML, XML, JSON, YAML) and found that format choice significantly impacts structural understanding tasks. He et al. [20] reported up to 40% performance variation based on prompt formatting alone. Chain-of-Table [22] demonstrated that LLMs can iteratively transform tables during reasoning. Our work extends this line of inquiry to agent-consumed schema files in file-native architectures.

TOON (Token-Oriented Object Notation) [10] is a compact serialisation format designed specifically for LLM context efficiency. It achieves approximately 25% smaller file sizes than YAML through abbreviated keywords (TABLE, COL, FK) and minimal syntax. We include TOON in our evaluation to test whether file size optimisation translates to runtime efficiency.

# 3 Methodology

## 3.1 Experimental Overview

**Table 2: Experimental Overview**

| Experiment | Focus | Evaluations | Key Variables |
|---|---|---|---|
| Core | SQL generation accuracy | 8,401 | Format, Model, Architecture, Tier |
| Scale | Schema navigation | 928 | Format, Scale tier (S0-S5) |
| Partition | Enterprise navigation | 320 | Partitioning approach (S6-S9) |

## 3.2 Format Conditions

We tested four schema representation formats: YAML (hierarchical structure, grep-friendly), Markdown (documentation-style with natural language), JSON (machine-parseable, verbose), and TOON (Token-Oriented Object Notation [10], a custom compact format ~25% smaller in file size). All formats received identical system prompts with generic grep guidance; no format-specific search patterns were provided. This tests native LLM comprehension of each format without optimisation. All formats include an identical navigator.md providing schema overview and table descriptions.

## 3.3 Architecture Conditions

We compared two context delivery architectures. File Agent: agents use grep and read tools to retrieve schema information from files on demand, selecting relevant sections based on query requirements. Prompt Baseline: the complete schema (~6,000 tokens for TPC-DS) is embedded in the system prompt alongside the query, with no tool access. Both conditions use identical schema content; only the delivery mechanism differs. This comparison isolates whether file-native retrieval provides advantages over direct context injection.

## 3.4 Model Conditions

We evaluated 11 models across three capability tiers.

**Table 3: Model Conditions**

| Tier | Models | Characteristics |
|---|---|---|
| Frontier | claude-opus-4.5, gpt-5.2, gemini-2.5-pro | Highest capability |
| Frontier Lab | claude-haiku-4.5, gpt-5-mini, gemini-2.5-flash | Cost-optimised from frontier labs |
| Open Source | DeepSeek-V3.2, kimi-k2, llama-4-maverick, llama-4-scout, qwen3-32b | Publicly available |

## 3.5 Complexity Tiers (L1-L5)

**Table 4: Query Complexity Tiers**

| Tier | Type | Tables | SQL Features |
|---|---|---|---|
| L1 | Direct lookup | 1 | SELECT, COUNT |
| L2 | Single join | 2 | Basic JOIN, aggregation |
| L3 | Multi-hop | 3-4 | Chained JOINs, GROUP BY |
| L4 | Complex aggregation | 4+ | CTEs, window functions |
| L5 | Multi-step reasoning | 5+ | Subqueries, nested logic |

## 3.6 Scale Tiers (S0-S9)

Scale tiers range from S0 (10 tables, startup MVP) through S5 (1,000 tables, large enterprise) for single-file schemas. Partitioned tiers S6-S9 extend to 10,000 tables using domain-based partitioning with ~250 tables per domain. Scale experiments used Claude models only.

## 3.7 Evaluation

Queries were derived from the TPC-DS benchmark [8] query patterns and curated to ensure coverage across L1-L5 complexity tiers. We use Jaccard similarity over result sets: $J(A,G) = |A \cap G| / |A \cup G|$, where A is the agent-generated result set and G is the ground truth. Success threshold: Jaccard >= 0.9. For multiple valid formulations, we credit any semantically correct approach. All models used temperature=0 for deterministic, reproducible outputs in a single-run design. The Claude Agent SDK, used for file-based Claude conditions, does not expose temperature control; OpenAI GPT-5 models enforce a minimum temperature of 1; all other conditions set temperature=0 explicitly via provider APIs. Statistical tests include independent t-tests, chi-square tests, and ANOVA with Benjamini-Hochberg correction [9].

# 4 Results

## 4.1 R1: File Agent vs Prompt Engineering

Finding: Architecture effectiveness depends on model tier.

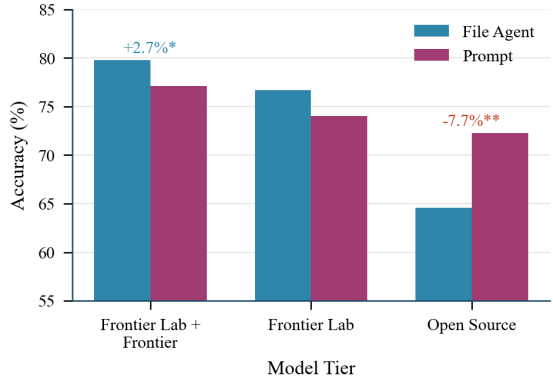

Figure 1: File Agent vs Prompt Engineering by Model Tier

Table 5: File Agent vs Prompt Engineering by Tier

| Model Tier | File Agent | Prompt | Diff | p-value | BH Sig |
|---|---|---|---|---|---|
| Frontier Lab + Frontier (excl. Opus) | 79.8% | 77.1% | +2.7% | 0.029 | Yes |
| Frontier Lab | 76.7% | 74.0% | +2.8% | 0.107 | No |
| Open Source | 64.6% | 72.3% | -7.7% | <0.001 | Yes |

The combined Frontier Lab + Frontier tier achieved significantly higher accuracy with file-native retrieval (+2.7%, p=0.029). Open source models showed mixed results: while the aggregate favoured prompt engineering by 7.7 percentage points (p<0.001), this was driven primarily by Qwen (-21.9%) and Llama Maverick (-13.9%), while Kimi (+0.3%) and Llama Scout (+0.5%) showed negligible differences.

Table 6: Individual Model File vs Prompt Comparison

| Model | Tier | File | Prompt | Diff | Winner |
|---|---|---|---|---|---|
| claude-opus-4.5 | Frontier | 89.0% | N/A | N/A | File only |
| gpt-5.2 | Frontier | 83.8% | 82.0% | +1.8% | File |
| gemini-2.5-pro | Frontier | 85.1% | 81.4% | +3.7% | File |
| claude-haiku-4.5 | Frontier Lab | 76.1% | 74.5% | +1.6% | File |
| gpt-5-mini | Frontier Lab | 77.1% | 69.4% | +7.7% | File |
| gemini-2.5-flash | Frontier Lab | 77.0% | 78.0% | -1.1% | Prompt |
| DeepSeek-V3.2 | Open Source | 75.3% | 78.6% | -3.3% | Prompt |
| kimi-k2 | Open Source | 75.0% | 74.7% | +0.3% | File |
| llama-4-maverick | Open Source | 60.9% | 74.8% | -13.9% | Prompt |
| llama-4-scout | Open Source | 63.1% | 62.6% | +0.5% | File |
| qwen3-32b | Open Source | 48.9% | 70.9% | -21.9% | Prompt |

File agent performed better for 6 models while prompt engineering performed better for 4 models. The largest file agent advantages appeared in frontier and frontier lab models (GPT-5-mini +7.7%, Gemini Pro +3.7%), while the largest prompt advantages appeared in open source models (Qwen -21.9%, Llama Maverick -13.9%).

## 4.2 R2: Format Effects on Accuracy

Finding: Format did not significantly affect aggregate accuracy.

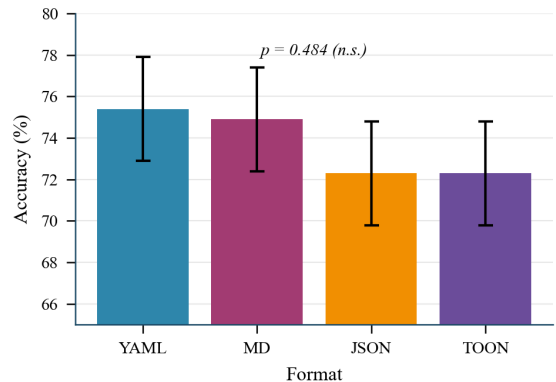

Figure 2: Accuracy by Format (File Agent)

Chi-squared test showed no significant format effect (p>0.05). YAML achieved 75.4%, MD 74.9%, JSON 72.3%, and TOON 72.3%. While aggregate results showed equivalence, individual models exhibited format sensitivities.

Table 7: Model × Format Accuracy (File Agent)

| Model | Tier | YAML | MD | JSON | TOON | Best | Spread |
|---|---|---|---|---|---|---|---|
| claude-opus-4.5 | Frontier | 92% | 86% | 88% | 90% | YAML | 5.4% |
| gemini-2.5-pro | Frontier | 86% | 84% | 88% | 83% | JSON | 5.1% |
| gpt-5.2 | Frontier | 84% | 83% | 85% | 83% | JSON | 1.8% |
| gpt-5-mini | Frontier Lab | 78% | 78% | 76% | 76% | MD | 1.6% |
| gemini-2.5-flash | Frontier Lab | 79% | 79% | 75% | 74% | YAML | 5.1% |
| claude-haiku-4.5 | Frontier Lab | 84% | 72% | 73% | 75% | YAML | 12.2% |
| DeepSeek-V3.2 | Open Source | 73% | 80% | 79% | 70% | MD | 10.2% |
| kimi-k2 | Open Source | 82% | 71% | 77% | 71% | YAML | 11.1% |
| llama-4-scout | Open Source | 60% | 72% | 59% | 61% | MD | 13.2% |
| llama-4-maverick | Open Source | 58% | 71% | 51% | 63% | MD | 20.1% |
| qwen3-32b | Open Source | 54% | 49% | 45% | 48% | YAML | 9.8% |

Format preference summary: YAML performed best for 5 models, MD for 4 models, JSON for 2 models, and TOON for 0 models. Open source models showed larger format sensitivities (spread 9.8-20.1%) than frontier models (spread 1.6-5.4%).

### 4.3 R3: Model Tier Effects

Finding: Model capability is the dominant factor.

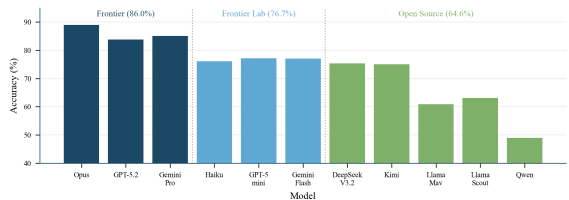

Figure 3: Accuracy by Individual Model

One-way ANOVA: $F(10, 8390) = 30.55$, $p < 0.001$. Frontier tier achieved 86.0% accuracy, Frontier Lab 76.7%, and Open Source 64.6%. The 21 percentage point gap between frontier and open source tiers dwarfs any format or architecture effect.

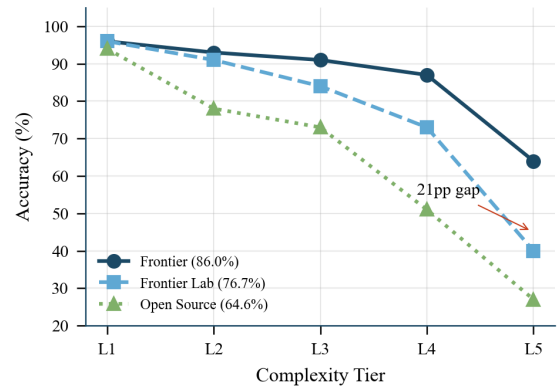

Figure 4: Accuracy by Complexity and Model Tier

All tiers achieved similar accuracy at L1 (94-96%), but diverged sharply at higher complexity. Frontier models maintained 64% at L5 while open source dropped to 27%.

Table 8: Individual Model Performance by Complexity (File Agent)

| Model | Tier | L1 | L2 | L3 | L4 | L5 | Overall |
|---|---|---|---|---|---|---|---|
| claude-opus-4.5 | Frontier | 100% | 93% | 94% | 94% | 64% | 89.0% |
| gemini-2.5-pro | Frontier | 94% | 97% | 81% | 81% | 74% | 85.1% |
| gpt-5.2 | Frontier | 95% | 89% | 96% | 85% | 54% | 83.8% |
| gpt-5-mini | Frontier Lab | 95% | 88% | 82% | 71% | 49% | 77.1% |
| gemini-2.5-flash | Frontier Lab | 94% | 92% | 83% | 75% | 40% | 77.0% |
| claude-haiku-4.5 | Frontier Lab | 100% | 93% | 87% | 71% | 29% | 76.1% |
| DeepSeek-V3.2 | Open Source | 96% | 92% | 82% | 67% | 40% | 75.3% |
| kimi-k2 | Open Source | 99% | 89% | 86% | 64% | 37% | 75.0% |
| llama-4-scout | Open Source | 94% | 75% | 79% | 50% | 18% | 63.1% |
| llama-4-maverick | Open Source | 90% | 82% | 73% | 37% | 23% | 60.9% |
| qwen3-32b | Open Source | 92% | 53% | 44% | 37% | 18% | 48.9% |

Notable observations: Gemini Pro achieved the highest L5 accuracy (74%) despite not having the highest overall score. Claude Opus maintained near-perfect L1-L3 performance. Qwen showed particularly steep degradation from L2 onwards.

### 4.4 R4: Navigation at Scale

Finding: File-native schema navigation scales to 10,000 tables with partitioning.

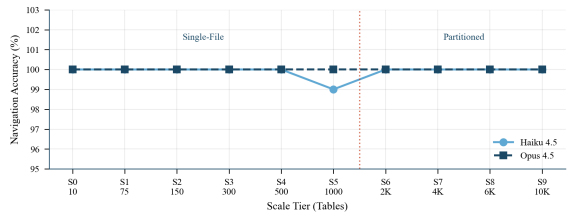

Figure 5: Navigation Accuracy at Scale

Single-file schemas maintained near-perfect accuracy up to 1,000 tables. Domain partitioning enabled high navigation accuracy at 10,000 tables. The partitioned architecture keeps per-query context bounded regardless of total schema size.

### 4.5 R5: Efficiency Effects

Finding: Token efficiency is model-dependent. Format efficiency varies significantly across models.

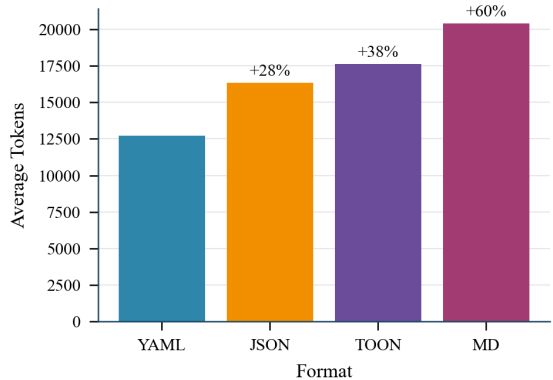

Figure 6: Token Efficiency by Format (TPC-DS, 11-Model Average)

Averaged across all 11 models on the TPC-DS schema (24 tables), YAML was the most token-efficient format (12,729 tokens), followed by JSON (16,320, +28%), TOON (17,625, +38%), and MD (20,382, +60%). Token efficiency varied substantially by model: Claude models consumed 935-2,579 tokens while open source models ranged from 3,692 to 37,841 tokens, reflecting differences in tool-use efficiency.

Markdown's higher token consumption in file-based agents was not caused by grep inefficiency. Models matched Markdown heading patterns effectively on first attempt. Rather, each grep result returned more tokens due to Markdown's verbose pipe-table formatting (column separators, alignment dashes, repeated delimiters) compared to YAML's compact key-value pairs. This overhead varied by model, ranging from +7% (Gemini Flash) to +105% (Claude Opus), with one outlier (Qwen 32B) where Markdown was actually 19% cheaper than YAML. Notably, this overhead was absent in prompt-based agents where the full schema was embedded directly, confirming the cause as retrieval verbosity rather than format comprehension.

TOON's higher token consumption had a different character to Markdown's. Despite being ~25% smaller in file size, TOON consumed 38% more tokens than YAML on average. Unlike Markdown, where each grep hit simply returned more text, TOON's overhead was driven by a combination of output density and additional tool calls from pattern unfamiliarity, discussed in Section 5.3.

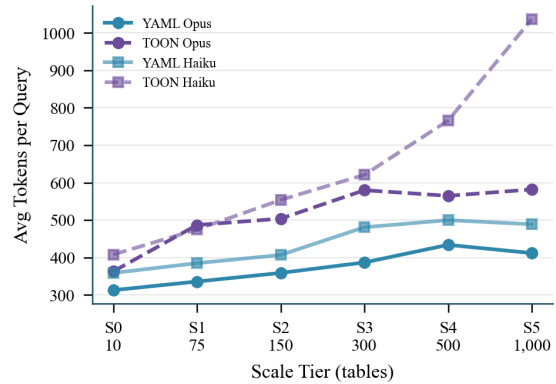

Figure 7: YAML vs TOON Token Usage

Scale experiments (Figure 7, Claude models only with the Agent SDK, N=20 per tier) revealed a model-dependent interaction. Both YAML lines and Opus TOON remained relatively stable across scale tiers, with Opus TOON maintaining a consistent ~1.37x ratio over YAML. Haiku TOON, however, diverged at larger schemas: from 408 tokens at S0 (1.14x) to 1,037 at S5 (2.12x), driven by increasing tool calls (1.6 at S0 to 5.0 at S5). The mechanism behind this divergence is discussed in Section 5.3.

## 5 Discussion

### 5.1 Architecture Choice is Model-Dependent

Our most significant finding challenges the assumption that file-native context retrieval universally improves agent performance. For frontier models, file-native retrieval provides measurable benefits (+2.7%, p=0.029). For open source models, results were mixed: some (Qwen, Llama Maverick) performed substantially worse with file-native retrieval, while others (Kimi, Llama Scout) showed no significant difference. We hypothesise this variation reflects differences in tool-use training across open source models, consistent with broader findings on capability gaps between open source and proprietary models [26].

### 5.2 Format Selection Should Be Operationally Driven

The absence of significant format effects on accuracy (p=0.484) suggests that format selection need not be driven by accuracy concerns. Instead, selection should be guided by: token efficiency (YAML consumed 28-60% fewer tokens across models), maintainability (YAML and JSON integrate with programmatic workflows), and grep-ability (formats with predictable patterns enable efficient retrieval).

### 5.3 The Grep Tax: Token Density and Pattern Unfamiliarity

The grep tax has two components. The first is token density: TOON's compact syntax packs more schema content per line, so each grep match returns denser output than YAML's spaced key-value structure. This overhead was visible across models in both the core experiment (24 tables, average TOON/YAML ratio 1.18x across all 11 models) and the scale experiment.

The second component is pattern unfamiliarity. TOON's custom keywords (TABLE, COL, FK) are unlikely to appear in current model training data. When grep returns partial matches, some models cycle through patterns drawn from familiar formats, trying DDL (CREATE TABLE), JSON ("TABLE_NAME":), and YAML (^ TABLE_NAME:) before discovering correct TOON syntax. In the worst observed case (S5_PK03, Haiku), 16 grep attempts were required versus 3 for the same query in YAML. Opus handled the same noisy output without excessive retries (max 4 tool calls at S5 vs Haiku's 16). By design, no format-specific prompt guidance was provided for grep patterns. Providing TOON-specific grep examples in the system prompt may mitigate this effect and is a candidate for future work.

The grep tax is also model-dependent. With the core experiment, models varied dramatically in TOON efficiency: DeepSeek consumed 10% fewer tokens with TOON than YAML, while Llama Scout consumed 38% more. This suggests different models have learned different grep patterns during training, with some effectively matching TOON's compact syntax.

This explains the divergence in Figure 7: Opus absorbs the token density baseline (~1.37x) without triggering pattern cycling, while Haiku experiences both components, with the noise-driven cycling becoming more frequent at larger schemas. Partition experiments using an index file approach with ~250 tables per domain showed no measurable increase in grep tax up to 10,000 tables, suggesting partitioning eliminates any scaling component. Scale experiments used Claude models and the Claude Agent SDK only; other models and SDKs may show different patterns.

### 5.4 Format Effects Are Model-Dependent

While format did not affect aggregate accuracy, we observed that formats influenced which valid interpretation models selected for semantically ambiguous queries. For queries about 'customer state', models took different join paths (home address via CUSTOMER vs transaction address via SS_ADDR_SK) depending on format. MD's flat structure guided 72% of models to one path while YAML's hierarchical structure guided only 33% to the same path. Notably, stronger models (Opus) navigated ambiguity correctly regardless of format, while weaker models followed format cues more readily. This suggests format structure subtly emphasises different relationships, but the effect only manifests where navigator guidance leaves room for interpretation. Practitioners should invest in navigator disambiguation rules rather than format optimisation.

### 5.5 Practical Recommendations

**Table 9: Architecture Selection Guide**

| Model Tier | Recommended Architecture |
|---|---|
| Frontier | File Agent |
| Frontier Lab | File Agent (validate first) |
| Open Source | Prompt Engineering |

**Table 10: Format Selection Guide**

| Priority | Recommendation |
|---|---|
| Token efficiency | YAML |
| Human readability | Markdown |
| Programmatic generation | YAML or JSON |
| Custom formats | Ensure grep-friendly patterns |

### 5.6 Limitations

The core experiment used a query bank of 100 queries (20 per tier); a larger query bank or different query construction methodology may yield different results. Scale experiments used Claude models only and tested schema navigation (metadata retrieval) rather than SQL reasoning at scale; SQL generation accuracy at 10,000 tables remains untested. All experiments used TPC-DS, a retail data warehouse benchmark. SQL generation serves as a proxy for programmatic operations but direct validation on non-SQL tasks would strengthen generalisability. TOON is a novel format largely absent from LLM training data; the observed grep tax may partly reflect format unfamiliarity by the model rather than an inherent limitation of compact formats. Format-specific prompt guidance could mitigate this effect, but was outside our methodology which tested native LLM comprehension without format-specific optimisation. Not all conditions enforced temperature=0: the Claude Agent SDK does not expose temperature control for file-based conditions, and GPT-5 models enforce a minimum temperature of 1; these conditions may exhibit variance not captured by our single-run design. Single-shot evaluation may understate accuracy achievable with multi-turn correction. File-based conditions used the Claude Agent SDK; grep and read tool implementations may vary between versions.

# 6 Conclusion

This paper presented a systematic study of context engineering for structured data, comprising 9,649 experiments across 11 models, 4 formats, 2 architectures, and schemas ranging from 10 to 10,000 tables.

Our key findings: (R1) Architecture effectiveness is model-dependent: file-native retrieval improved accuracy for frontier models but showed mixed results for open source models. (R2) Format does not significantly affect accuracy at the aggregate level, but individual models, particularly open source, show meaningful format sensitivities. (R3) Model capability is the dominant factor, with a 21-percentage point gap between tiers. (R4) File-native schema navigation scales to 10,000 tables with domain partitioning. (R5) Token efficiency varies by format: compact or novel formats can incur a 'grep tax' driven by output density and pattern unfamiliarity, with the magnitude depending on model capability.

For practitioners: match architecture to model capability rather than assuming universal best practices. Invest in model capability before optimising format. Use YAML for token efficiency and grep-friendly patterns. For novel or compact formats, use prompt guidance to support grep behaviours and design context engineering to reduce unnecessary token density in grep results. Partition schemas by domain for enterprise scale.

Future work should extend experiments to non-SQL tasks, validate scale findings across providers, investigate multi-turn evaluation, and test whether compact formats can be optmised for efficient grep-ability. As LLM agents increasingly operate critical business systems, evidence-based guidance on context engineering becomes essential.